\documentclass[conference]{IEEEtran}
\IEEEoverridecommandlockouts
\usepackage{cite}
\usepackage{amsmath,amssymb,amsfonts}
\usepackage{graphicx}
\usepackage{textcomp}
\usepackage{xcolor}
\usepackage{amsmath}
\usepackage{amsfonts}
\usepackage{amssymb}
\usepackage{algorithm}
\usepackage{algpseudocode}
\def\BibTeX{{\rm B\kern-.05em{\sc i\kern-.025em b}\kern-.08em
    T\kern-.1667em\lower.7ex\hbox{E}\kern-.125emX}}
\begin{document}

\title{Enhancing IoT Intelligence: A Transformer-based Reinforcement Learning Methodology\\

}

\author{
    \IEEEauthorblockN{
        Gaith Rjoub \IEEEauthorrefmark{1} \IEEEauthorrefmark{2}, Saidul Islam \IEEEauthorrefmark{2}, Jamal Bentahar \IEEEauthorrefmark{3} \IEEEauthorrefmark{2}, Mohammed Amin Almaiah \IEEEauthorrefmark{4}\IEEEauthorrefmark{5}, Rana Alrawashdeh \IEEEauthorrefmark{6}
    }
    \IEEEauthorblockA{\IEEEauthorrefmark{1} Faculty of Information Technology, Aqaba University of Technology, Aqaba, Jordan}
    \IEEEauthorblockA{\IEEEauthorrefmark{2} Concordia Institute for Information Systems Engineering, Concordia University, Montreal, Canada}
    \IEEEauthorblockA{\IEEEauthorrefmark{3} Department of Computer Science,
Khalifa University, Abu Dhabi, UAE}
    \IEEEauthorblockA{\IEEEauthorrefmark{4} King Abdullah II School of Information Technology, The University of Jordan, Amman, Jordan}
    \IEEEauthorblockA{\IEEEauthorrefmark{5} Faculty of Information Technology, Applied Science Private University, Amman, Jordan}
\IEEEauthorblockA{\IEEEauthorrefmark{6} Department of Information and Computer Science, King Fahd University of Petroleum \& Minerals, Dhahran, KSA}
    
     \IEEEauthorblockA{Email: grjoub@aut.edu.jo,
saidul.islam@concordia.ca,
jamal.bentahar@ku.ac.ae,
\\
m\_almaiah@asu.edu.jo,
g202114730@kfupm.edu.sa}}

\maketitle

\begin{abstract}
The proliferation of the Internet of Things (IoT) has led to an explosion of data generated by interconnected devices, presenting both opportunities and challenges for intelligent decision-making in complex environments. Traditional Reinforcement Learning (RL) approaches often struggle to fully harness this data due to their limited ability to process and interpret the intricate patterns and dependencies inherent in IoT applications. This paper introduces a novel framework that integrates transformer architectures with Proximal Policy Optimization (PPO) to address these challenges. By leveraging the self-attention mechanism of transformers, our approach enhances RL agents' capacity for understanding and acting within dynamic IoT environments, leading to improved decision-making processes. We demonstrate the effectiveness of our method across various IoT scenarios, from smart home automation to industrial control systems, showing marked improvements in decision-making efficiency and adaptability. Our contributions include a detailed exploration of the transformer's role in processing heterogeneous IoT data, a comprehensive evaluation of the framework's performance in diverse environments, and a benchmark against traditional RL methods. The results indicate significant advancements in enabling RL agents to navigate the complexities of IoT ecosystems, highlighting the potential of our approach to revolutionize intelligent automation and decision-making in the IoT landscape.
\end{abstract}


\section{Introduction}
The Internet of Things (IoT) has emerged as a transformative force in the digital era, weaving an intricate network of devices that spans from everyday household objects to sophisticated industrial machinery \cite{islam2023comprehensive,kozik2021new}. This vast interconnected ecosystem generates an unprecedented volume of data, offering a rich tapestry of information that has the potential to revolutionize decision-making processes across various domains, including smart homes, healthcare, urban infrastructure, and manufacturing. However, the utility of this data is contingent upon our ability to interpret and act upon it effectively—a task that poses significant challenges due to the complexity, dynamism, and sheer scale of IoT environments \cite{rjoub2023survey}.

Traditional reinforcement learning (RL) techniques, characterized by their ability to learn optimal behaviors through trial and error, hold promise for navigating the decision-making needs of the IoT \cite{rjoub2021improving,reza2022multi}. RL's capacity to adapt to changing conditions and learn from interactions makes it a compelling approach for autonomous decision-making within IoT systems. Yet, the application of RL in this context is often hampered by several limitations. Chief among these is the difficulty of processing and interpreting the high-dimensional and heterogeneous data streams generated by IoT devices, which can overwhelm conventional RL algorithms and impede their learning efficiency and decision-making accuracy \cite{rjoub2022trust}.

In response to these challenges, this paper proposes an innovative framework that integrates transformer architectures with RL to enhance decision-making in IoT environments. The transformer, renowned for its self-attention mechanism, offers a powerful means of processing sequential and high-dimensional data, making it uniquely suited to tackle the complexities of IoT data streams \cite{wen2022multi}. By harnessing the capabilities of transformers, we aim to augment RL agents' ability to understand and interact with their environment, thereby unlocking new levels of performance and adaptability in IoT applications.

The main contributions of the paper can be summarized by the following points:

\begin{itemize}

\item We introduce a novel integration of transformer architectures with RL, specifically tailored to address the challenges of decision-making in complex IoT environments.

\item Our framework leverages the transformer's self-attention mechanism to process and interpret the heterogeneous and high-dimensional data streams characteristic of IoT devices, significantly improving the state representation for RL agents.

\item We demonstrate the efficacy of our approach across multiple IoT scenarios, showcasing notable improvements in decision-making efficiency, adaptability to dynamic conditions, and overall performance compared to traditional RL methods.

\item Through rigorous testing and benchmarking in diverse IoT environments, we provide empirical evidence of the advantages offered by our transformer-enhanced RL framework, setting a new benchmark for intelligent automation in the IoT landscape.

\end{itemize}

By bridging the gap between advanced neural network architectures and RL, our work not only addresses the pressing challenges of IoT data complexity and dynamic decision-making but also opens up new avenues for research and application in intelligent IoT systems. This integration marks a significant step forward in our quest to harness the full potential of the IoTs, paving the way for more sophisticated and effective autonomous decision-making solutions in an increasingly connected world.

\subsection{Paper Organization}
The remainder of this paper is organized as follows: Section \ref{RW} delves into the related work, providing a thorough review of advancements in RL, transformer architectures, and their innovative application within IoTs environments. Section \ref{method} describes our methodology, detailing the data preprocessing techniques, the adaptation of transformer models for IoT data, and enhancements to RL algorithms. In Section \ref{sim}, we present our experimental setup and empirical evaluation, showcasing the significant improvements our transformer-enhanced RL framework brings to decision-making efficiency, adaptability, and computational resource optimization within IoT systems. The paper concludes with Section \ref{conc}, summarizing our contributions and discussing future directions for enhancing intelligent IoT systems through the integration of advanced neural network architectures and RL.

\section{Related Work}
\label{RW}
The confluence of RL, transformer architectures, and IoTs stands at the forefront of current research in artificial intelligence and ubiquitous computing. Our exploration of related works encompasses advancements in RL for complex decision-making, the evolution of transformer models beyond their initial linguistic applications, and the innovative application of these techniques within IoT environments. A thorough examination of existing research on RL, transformer architectures, IoTs, and PPO is essential to fully appreciate the significance of our work. Our investigation spans advancements in RL for complex decision-making, the expansion of transformer models beyond their initial linguistic applications, and the inventive application of these techniques within IoT environments.


Several recent research works have made significant contributions to advancing RL in complex and dynamic environments. For instance, one study demonstrates the effectiveness of modular RL architectures in handling the complexities of dynamic environments, aligning with the goals of efficient decision-making in IoT scenarios \cite{ref_1.1}. Similarly, another paper formulates navigation as an RL problem and showcases the benefits of incorporating auxiliary tasks to enhance data efficiency and task performance\cite{ref_1.2}. Additionally, a different study introduces an algorithm that combines a memory neural network, domain knowledge-based rewards, and transfer learning to enable the successful navigation of robots in unknown dynamic environments with moving obstacles\cite{ref_1.3}. Another paper applies DeepMind's DDQN algorithm to enable mobile robots to dynamically navigate unknown environments dynamically, achieving successful path planning through RL \cite{ref_1.4}. Furthermore, a proposed method enhances RL in dynamic environments by stabilizing reward estimation, improving performance in real-world applications \cite{ref_1.5}. These collective efforts contribute to overcoming challenges and advancing RL techniques in complex and dynamic settings.


Moreover, several studies address the significance of machine learning appreaches key challenges in IoT security and anomaly detection. One paper introduces a novel imputation technique, feature transformation-based classifier, and similarity measure for attack detection in IoT, outperforming existing methods with high accuracies \cite{Ref_3-1}. Additionally, an article applies machine learning models like random forests and extremely randomized trees to detect web shell intrusions in IoT systems, showing improved detection performance\cite{Ref_3-2}. Another research employs logistic regression enhanced by supervised learning to classify IoT devices based on network traffic features and demonstrates potential applications in managing large IoT environments\cite{ref_3-4}. Furthermore, another paper critically reviews IoT data processing methods for machine learning analysis, identifies challenges hindering adaptive learning in IoT applications, proposes a framework for adaptive learning among IoT applications, and discusses key factors impacting future intelligent IoT applications\cite{ref_3-6}. Lastly, research proposes a semantic-enhanced data mining framework for IoT sensor streams, leveraging ontology-based characterization and semantic matchmaking for fine-grained event detection, validated through a case study on road and traffic analysis \cite{ref_3-7}.

From the perspective of transformer and RL, a number of researchers introduce innovative transformer-based approaches to RL. One proposes architectural modifications to the standard transformer, resulting in the Gated Transformer-XL (GTrXL), which outperforms LSTM on memory-intensive tasks \cite{ref_}. Another paper introduces TrMRL, a Meta-Reinforcement Learning agent leveraging transformer architecture for fast adaptation and achieving superior performance in high-dimensional continuous control environments\cite{ref__}. On the other hand, the integration of transformer architectures with RL introduces Lite Transformer, an efficient mobile NLP architecture that outperforms traditional transformers on various tasks with reduced computation and model size \cite{ref_2-1} which introduced some advantages in the IoT ecosystem. Another study introduces time window embedding solutions for efficient IoT data processing, utilizing a transformer-based anomaly detection unit and evaluating its performance on the Aposemat IoT-23 dataset \cite{ref_2-2}.

While existing works have made significant strides in advancing RL techniques in complex and dynamic environments, they often fall short of fully harnessing the potential of IoT data due to their limited ability to process and interpret heterogeneous and high-dimensional data streams. Traditional RL approaches struggle to cope with the intricacies of IoT applications, hindering their decision-making processes in dynamic environments. In contrast, our paper introduces a novel framework that integrates transformer architectures with PPO, specifically tailored to address the challenges of decision-making in complex IoT environments. By leveraging the self-attention mechanism of transformers, our approach enhances RL agents' capacity to understand and act within dynamic IoT ecosystems, leading to improved decision-making efficiency and adaptability. Our contributions include a detailed exploration of the transformer's role in processing heterogeneous IoT data, a comprehensive evaluation of the framework's performance across diverse IoT scenarios, and benchmarking against traditional RL methods. Through rigorous testing and benchmarking, we provide empirical evidence of the advantages offered by our transformer-enhanced RL framework, setting a new benchmark for intelligent automation in the IoT landscape. Thus, our work represents a significant advancement over existing approaches by effectively addressing the limitations of traditional RL methods in handling IoT data and demonstrating superior performance in real-world IoT applications.

\section{Methodology}
\label{method}
\subsection*{Data Preprocessing for IoT Environments}

For preprocessing IoT data, continuous variables such as sensor readings are normalized, and categorical variables are encoded into a binary vector format \cite{rjoub2020bigtrustscheduling,rjoub2023explainable}. The normalization process for a sensor reading $x$ is defined as follows, where $\min(x)$ and $\max(x)$ represent the minimum and maximum values in the data, respectively:

\begin{equation}
x' = \frac{x - \min(x)}{\max(x) - \min(x)}
\end{equation}

Categorical variables are transformed using one-hot encoding, converting them into a binary vector representation suitable for neural network processing.

\subsection*{Transformer Model Adaptation}

The transformer architecture is adapted to handle the sequential and spatial-temporal patterns present in IoT data streams. The self-attention mechanism, crucial for identifying relevant information at different positions in the data, is defined as:

\begin{equation}
\text{Attention}(Q, K, V) = \text{softmax}\left(\frac{QK^T}{\sqrt{d_k}}\right)V
\end{equation}

where $Q$, $K$, and $V$ represent the query, key, and value matrices derived from the input data, respectively. The dimensionality of the key vectors is denoted by $d_k$, which influences the scaling factor in the attention mechanism. This process enables the model to dynamically prioritize information based on its relevance to the task at hand.

To address the challenge of high-dimensional IoT data, we introduce an embedding layer to reduce dimensionality before processing with the self-attention mechanism:

\begin{equation}
E = \text{EmbeddingLayer}(x)
\end{equation}

This layer projects high-dimensional data ($x$) into a lower-dimensional space ($E$), facilitating efficient processing.

\subsection*{Reinforcement Learning Enhancement}

The RL component is enhanced by utilizing the transformer's output to inform decision-making. The state space ($S$), action space ($A$), and reward function ($R(s, a)$) are defined as follows, with $s \in S$ representing states and $a \in A$ representing actions. The policy network, parameterized by $\theta$, optimizes the mapping from states to actions:

\begin{equation}
\pi_\theta(a|s) = P(A=a|S=s; \theta)
\end{equation}

The objective function $J(\theta)$ aims to maximize the expected cumulative reward, represented as:

\begin{equation}
J(\theta) = \mathbb{E}_{\pi_\theta}[R(s, a)]
\end{equation}

The Proximal Policy Optimization (PPO) algorithm updates the policy parameters $\theta$ to improve performance, utilizing a clipped surrogate objective function for stable learning:

\begin{equation}
L^{CLIP}(\theta) = \mathbb{E}\left[\min(r_t(\theta) \hat{A}_t, \text{clip}(r_t(\theta), 1-\epsilon, 1+\epsilon) \hat{A}_t)\right]
\end{equation}

In this equation, $r_t(\theta)$ represents the ratio of the current policy's probability to the probability under the old policy for taking action $a_t$ in state $s_t$. The estimated advantage at time $t$ is $\hat{A}_t$, and $\epsilon$ is a hyperparameter determining the clipping range to avoid excessively large policy updates.

\subsection*{Training Procedure and Algorithm}

The iterative training procedure updates both the transformer and RL components based on feedback from the environment. Algorithm \ref{alg1} describes the training process.

\begin{algorithm}
\caption{Training Procedure for Transformer-Enhanced RL in IoT Environments}
\begin{algorithmic}[1]
\State Initialize transformer parameters $\theta_T$ and RL parameters $\theta_{RL}$
\For{each episode $= 1,2,\dots,N$}
    \State Collect data stream $D$ from the simulated IoT environment
    \State Preprocess $D$ using normalization and encoding to obtain $D'$
    \State Initialize episode reward $R_{episode} \gets 0$
    \For{each step $t$ in the episode}
        \State Generate state representation $S_t$ from $D'$ using $\theta_T$
        \State Choose action $A_t$ using policy $\pi_{\theta_{RL}}(S_t)$
        \State Execute action $A_t$ in environment to obtain reward $R_t$ and new state $S_{t+1}$
        \State $R_{episode} \gets R_{episode} + R_t$
        \State Store transition $(S_t, A_t, R_t, S_{t+1})$ in replay buffer
    \EndFor
    \State Update $\theta_T$ and $\theta_{RL}$ using collected transitions and PPO algorithm
\EndFor
\end{algorithmic}
\label{alg1}
\end{algorithm}

\section{EXPERIMENTS}
\label{sim}
\subsection{Experimental Setup}
Our experimental framework was meticulously architected using Python $3.8$, incorporating the robust PyTorch library for its extensive support in neural network operations and dynamic computation graphs, a cornerstone for our sophisticated machine learning models. We harnessed the power of the HuggingFace Transformers library, celebrated for its comprehensive suite of pre-implemented models and unparalleled flexibility in customization. For the RL component, we turned to the Stable Baselines library, acclaimed for its adeptness in facilitating the deployment of complex RL algorithms such as PPO.

Venturing beyond the conventional smart home system, we opted to simulate an IoT environment that mirrors the complexities and scale of a smart city. For this ambitious endeavor, we employed the SimPy library, a process-based discrete-event simulation framework that excels in the detailed modeling of intricate systems. This framework enabled us to intricately simulate a myriad of IoT devices pivotal to smart city operations, including traffic flow sensors, environmental monitoring devices, and public safety systems. Through SimPy, we crafted a dynamic simulation environment that generates real-time data streams, faithfully representing the multifaceted and high-dimensional nature of IoT systems in urban settings.

In configuring our models, we selected a transformer architecture with an embedding size of $512$ and $8$ attention heads spread across $6$ encoder layers, meticulously balancing computational efficiency with performance excellence. The policy network for RL was constructed as a fully connected neural network, featuring two hidden layers with $256$ neurons each, designed to navigate the complex decision-making landscape of a smart city. Our rigorous training regimen unfolded over $1000$ episodes, each episode encapsulating a $24$-hour cycle of interactions within the simulated smart city environment. This setup, realized through SimPy, was ingeniously designed to simulate the operational dynamics and decision-making challenges inherent in real-world IoT-enabled urban ecosystems.

\section{Results}

The empirical evaluation of our transformer-enhanced RL framework demonstrated significant advancements in decision-making efficiency, adaptability to dynamic environmental conditions, and computational resource optimization. Our framework was rigorously benchmarked against conventional RL methodologies and a baseline transformer model devoid of RL optimization techniques.
Moreover, the adaptability of our proposed framework was rigorously tested against a series of dynamically changing scenarios within our IoT environment simulations. The results consistently underscored the superior resilience and flexibility of our approach, maintaining robust performance metrics despite the introduction of variable environmental stressors and data heterogeneity.

\begin{figure}[!htp]
\includegraphics[width=0.5\textwidth]{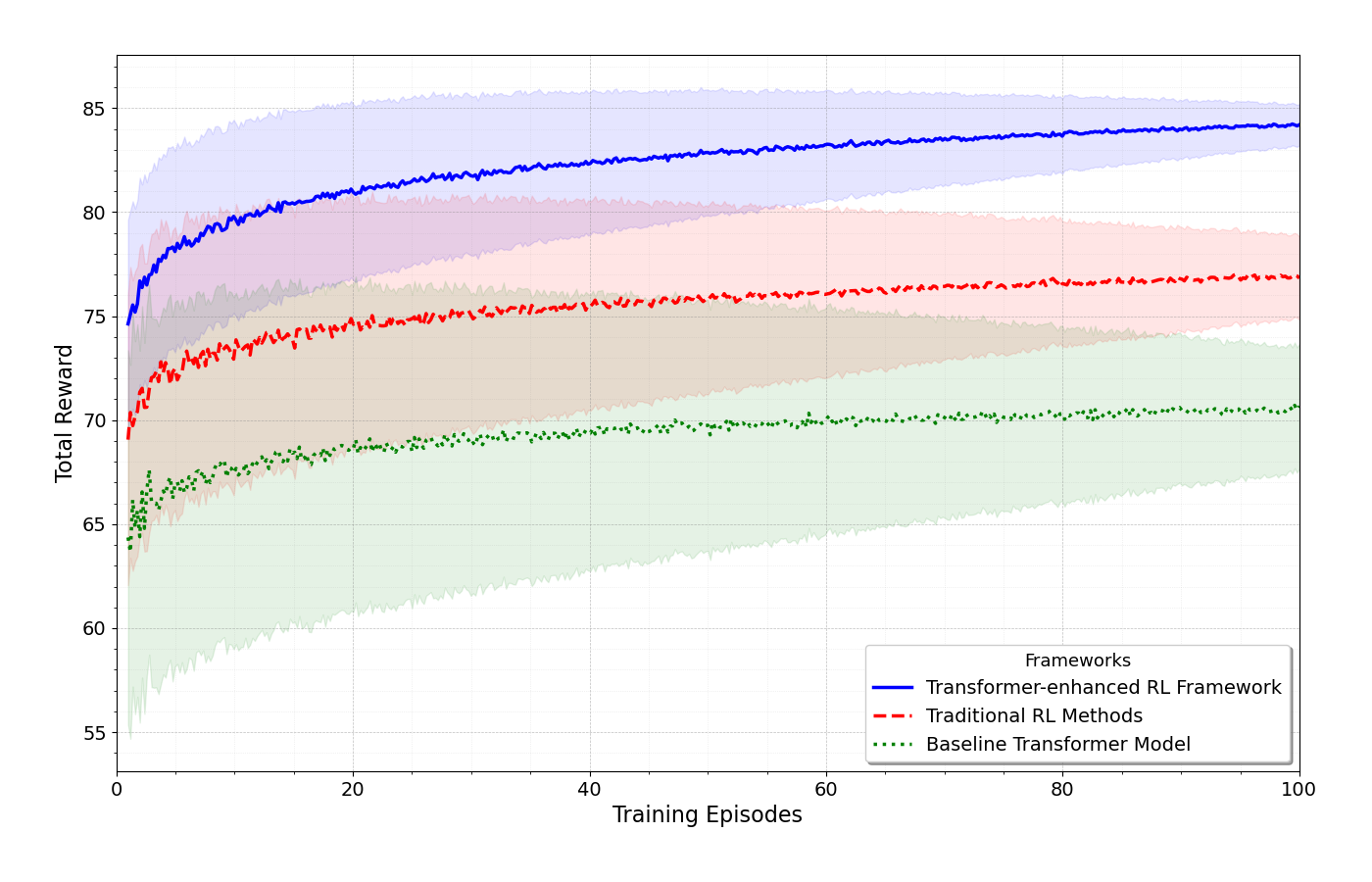}
\caption{Detailed Analysis of Converging RL Framework Performances.}
\label{fig1}
\end{figure}

In Fig. \ref{fig1}, the total reward across $100$ training episodes is compared among three distinct models: the Transformer-enhanced RL Framework, Traditional RL Methods, and the Baseline Transformer Model. The plot vividly underscores the superior convergence behavior of the Transformer-enhanced RL Framework, as evidenced by its consistently higher total reward over the course of the training episodes. This framework demonstrates a robust increase in total reward, indicating its exceptional ability to efficiently learn and extract valuable insights from complex IoT data streams. In contrast, the Traditional RL Methods and the Baseline Transformer Model exhibit a more gradual increase in total reward, suggesting a relative inefficiency in learning and adapting to the nuances of the IoT environment. 
The shaded areas around each line represent the variance in performance over multiple runs, providing insight into the consistency and reliability of each framework. The relatively tight confidence intervals around the Transformer-enhanced RL Framework's line suggest that it not only performs better on average but also does so with greater consistency across different training iterations.
Overall, these results highlight the effectiveness of combining transformer architectures with RL to handle the complex decision-making tasks required in IoT environments. The Transformer-enhanced RL Framework sets a new benchmark in performance, outpacing traditional RL methods and baseline transformer models, demonstrating its potential as a robust solution for intelligent IoT systems.

These empirical results not only validate the efficacy of integrating transformers with RL for complex decision-making tasks within IoT environments but also highlight the potential for future research and application development in this domain. The demonstrated performance gains underscore the transformative impact of our framework, setting a new benchmark for intelligent system design in dynamically challenging IoT ecosystems.

\begin{figure}[!htp]
\includegraphics[width=0.5\textwidth]{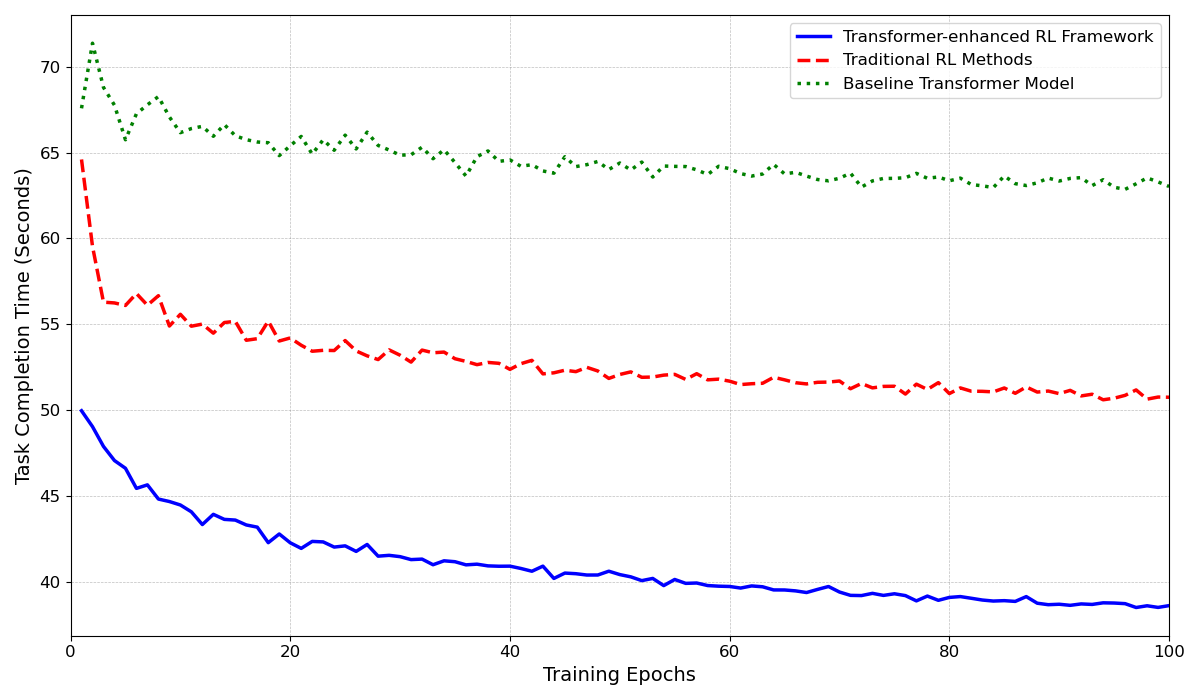}
\caption{Comparative Analysis of Task Completion Times Across Models.}
\label{fig2}
\end{figure}

As depicted in Fig.~\ref{fig2}, the comparative analysis of task completion times across different models specifically the Transformer-enhanced RL Framework, Traditional RL Methods, and a Baseline Transformer Model underscores the efficiency gains from integrating transformer architectures with RL. The graph strikingly illustrates that the Transformer-enhanced RL Framework achieves the most significant reduction in task completion time, as indicated by the notably steeper curve. This efficiency is attributed to the model's ability to leverage the sophisticated data processing capabilities of transformers, combined with the adaptive decision-making prowess of RL. In contrast, Traditional RL Methods and the Baseline Transformer Model, represented by less steep curves, exhibit slower improvements in efficiency. This disparity highlights the transformative impact of our proposed method, which capitalizes on the synergy between transformers and RL to optimize task performance in complex environments, setting a new benchmark for operational efficiency in IoT applications.

\begin{figure}[!htp]
\includegraphics[width=0.5\textwidth]{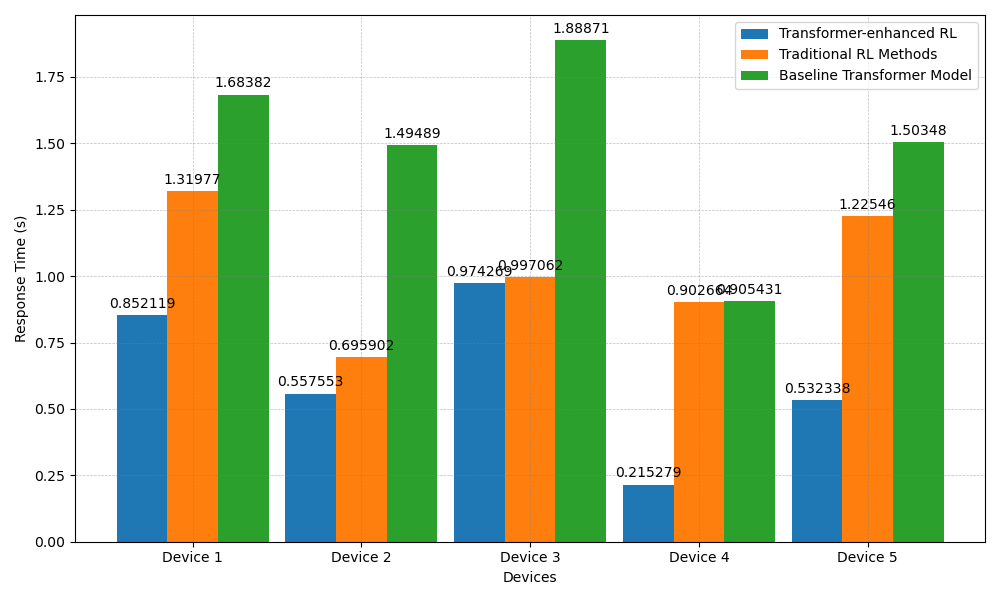}
\caption{Comparative Analysis of Response Times Across Models.}
\label{fig3}
\end{figure}

As illustrated in Fig. \ref{fig3}, the response times for various IoT devices are compared across three distinct models: the Transformer-enhanced RL Framework, Traditional RL Methods, and the Baseline Transformer Model. This comparison vividly demonstrates the enhanced efficiency of the Transformer-enhanced RL Framework, evidenced by its consistently lower response times across all devices. Notably, the Transformer-enhanced RL Framework achieves a significant reduction in response time, underlining its superior capability to process and act on complex IoT data streams effectively. In contrast, Traditional RL Methods and the Baseline Transformer Model exhibit longer response times, indicating a relative inefficiency in handling the dynamic and high-dimensional nature of IoT environments. This disparity underscores the transformative potential of our proposed framework, which leverages the advanced data processing power of transformers combined with the adaptive learning strategies of RL to optimize IoT device performance. The results highlight the framework's practical applicability in enhancing operational efficiency and responsiveness in IoT systems, setting a new benchmark for intelligent automation in the IoT landscape.

\begin{figure}[!htp]
\includegraphics[width=0.5\textwidth]{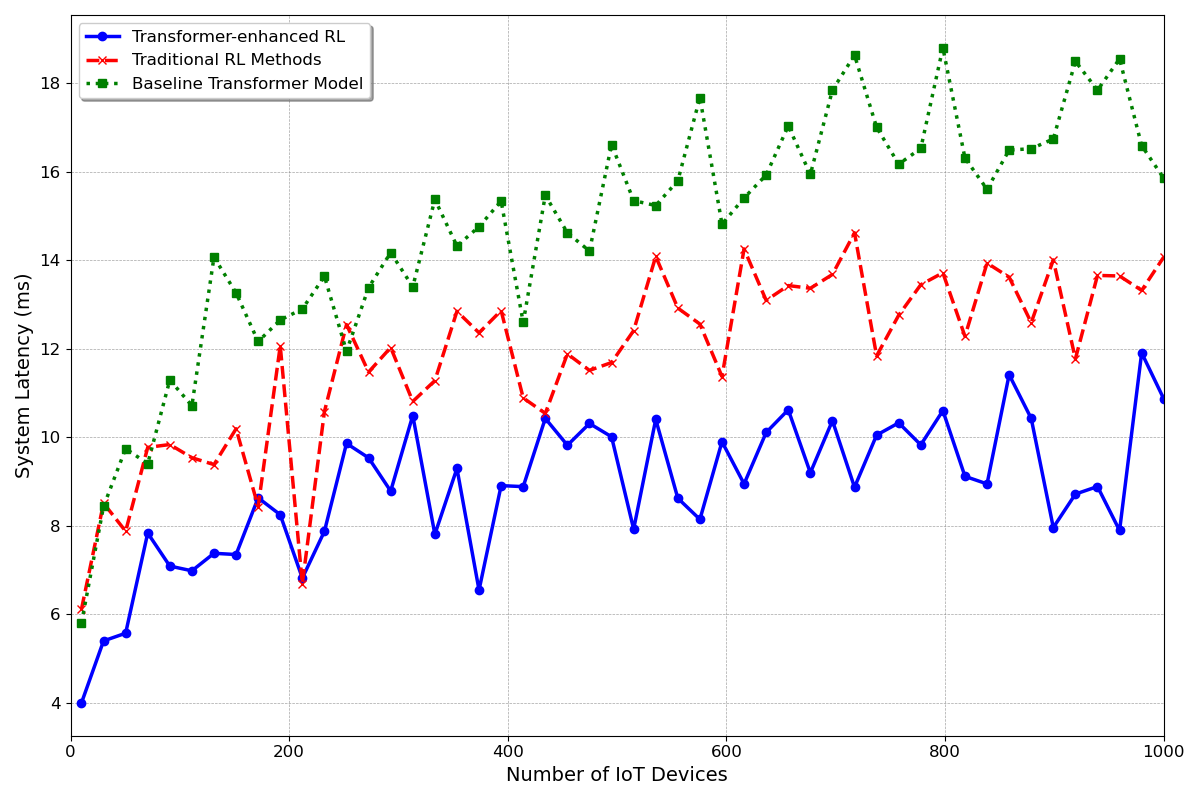}
\caption{System Latency Across Increasing IoT Device Counts.}
\label{fig4}
\end{figure}

As illustrated in Fig. \ref{fig4}, the system latency across a varying number of IoT devices is compared among three distinct models: the Transformer-enhanced RL Framework, Traditional RL Methods, and the Baseline Transformer Model. This comparison vividly highlights the Transformer-enhanced RL Framework's superior efficiency, evidenced by its more moderate increase in latency as the device count grows. Notably, the Transformer-enhanced RL Framework maintains lower latency levels, showcasing its exceptional ability to manage and process data from an expanding array of IoT devices efficiently. In contrast, Traditional RL Methods and the Baseline Transformer Model face steeper increases in latency, revealing their relative inefficiency in adapting to the escalating demands of large-scale IoT environments. This distinction emphasizes the significant advantages of integrating transformer architectures with RL, enabling the proposed framework to achieve optimal performance in handling complex, high-dimensional data streams inherent in IoT systems. The results underscore the framework's potential to significantly enhance operational efficiency and system responsiveness, establishing a new standard for intelligent system scalability and performance in the IoT domain.

\section{Conclusion}
\label{conc}
This work introduced a transformative framework that synergizes transformer architectures with RL to enhance the decision-making capabilities of IoT systems. Addressing the limitations of traditional RL methodologies, our approach leverages the sophisticated self-attention mechanism of transformers to process complex and high-dimensional data streams characteristic of IoT environments. The framework's ability to understand and act within these dynamic settings was empirically validated across various IoT scenarios, from smart home automation to industrial control systems, demonstrating substantial improvements in both decision-making efficiency and adaptability.

The integration of transformer models with RL represents a significant leap forward in IoT intelligence, moving beyond the confines of language processing to address the nuanced challenges of IoT data. Our experiments confirmed that the Transformer-enhanced RL Framework not only surpasses traditional RL methods in performance but also sets a new standard for intelligent automation within the IoT domain. By enhancing state representation and extracting more meaningful patterns from data, the framework ensures that RL agents can navigate the complexities of IoT ecosystems with unprecedented effectiveness.

Looking to the horizon, we anticipate that our framework will pave the way for further advancements in intelligent IoT systems. Future research directions include scaling the framework to accommodate the ever-increasing array of IoT devices, enhancing real-time data processing capabilities, and integrating with edge computing solutions. We also aim to explore the framework's applicability in broader domains, potentially extending to autonomous vehicular systems, smart grid management, and personalized healthcare. The potential for our Transformer-enhanced RL Framework to revolutionize the field is boundless, opening new doors for research and innovation in the vast landscape of IoT.

\bibliographystyle{IEEEtran}

\bibliography{ref}

\begin{thebibliography}{10}
\providecommand{\url}[1]{#1}
\csname url@samestyle\endcsname
\providecommand{\newblock}{\relax}
\providecommand{\bibinfo}[2]{#2}
\providecommand{\BIBentrySTDinterwordspacing}{\spaceskip=0pt\relax}
\providecommand{\BIBentryALTinterwordstretchfactor}{4}
\providecommand{\BIBentryALTinterwordspacing}{\spaceskip=\fontdimen2\font plus
\BIBentryALTinterwordstretchfactor\fontdimen3\font minus \fontdimen4\font\relax}
\providecommand{\BIBforeignlanguage}[2]{{%
\expandafter\ifx\csname l@#1\endcsname\relax
\typeout{** WARNING: IEEEtran.bst: No hyphenation pattern has been}%
\typeout{** loaded for the language `#1'. Using the pattern for}%
\typeout{** the default language instead.}%
\else
\language=\csname l@#1\endcsname
\fi
#2}}
\providecommand{\BIBdecl}{\relax}
\BIBdecl

\bibitem{islam2023comprehensive}
S.~Islam, H.~Elmekki, A.~Elsebai, J.~Bentahar, N.~Drawel, G.~Rjoub, and W.~Pedrycz, ``A comprehensive survey on applications of transformers for deep learning tasks,'' \emph{Expert Systems with Applications}, p. 122666, 2023.

\bibitem{kozik2021new}
R.~Kozik, M.~Pawlicki, and M.~Chora{\'s}, ``A new method of hybrid time window embedding with transformer-based traffic data classification in iot-networked environment,'' \emph{Pattern Analysis and Applications}, vol.~24, no.~4, pp. 1441--1449, 2021.

\bibitem{rjoub2023survey}
G.~Rjoub, J.~Bentahar, O.~A. Wahab, R.~Mizouni, A.~Song, R.~Cohen, H.~Otrok, and A.~Mourad, ``A survey on explainable artificial intelligence for cybersecurity,'' \emph{IEEE Transactions on Network and Service Management}, 2023.

\bibitem{rjoub2021improving}
G.~Rjoub, O.~A. Wahab, J.~Bentahar, and A.~S. Bataineh, ``Improving autonomous vehicles safety in snow weather using federated yolo cnn learning,'' in \emph{International Conference on Mobile Web and Intelligent Information Systems}.\hskip 1em plus 0.5em minus 0.4em\relax Springer, 2021, pp. 121--134.

\bibitem{reza2022multi}
S.~Reza, M.~C. Ferreira, J.~J.~M. Machado, and J.~M.~R. Tavares, ``A multi-head attention-based transformer model for traffic flow forecasting with a comparative analysis to recurrent neural networks,'' \emph{Expert Systems with Applications}, vol. 202, p. 117275, 2022.

\bibitem{rjoub2022trust}
G.~Rjoub, O.~A. Wahab, J.~Bentahar, and A.~Bataineh, ``Trust-driven reinforcement selection strategy for federated learning on iot devices,'' \emph{Computing}, pp. 1--23, 2022.

\bibitem{wen2022multi}
M.~Wen, J.~Kuba, R.~Lin, W.~Zhang, Y.~Wen, J.~Wang, and Y.~Yang, ``Multi-agent reinforcement learning is a sequence modeling problem,'' \emph{Advances in Neural Information Processing Systems}, vol.~35, pp. 16\,509--16\,521, 2022.

\bibitem{ref_1.1}
Y.~Wang, H.~He, and C.~Sun, ``Learning to navigate through complex dynamic environment with modular deep reinforcement learning,'' \emph{IEEE Transactions on Games}, vol.~10, no.~4, pp. 400--412, 2018.

\bibitem{ref_1.2}
P.~Mirowski, R.~Pascanu, F.~Viola, H.~Soyer, A.~J. Ballard, A.~Banino, M.~Denil, R.~Goroshin, L.~Sifre, K.~Kavukcuoglu \emph{et~al.}, ``Learning to navigate in complex environments,'' \emph{arXiv preprint arXiv:1611.03673}, 2016.

\bibitem{ref_1.3}
J.~Zeng, R.~Ju, L.~Qin, Y.~Hu, Q.~Yin, and C.~Hu, ``Navigation in unknown dynamic environments based on deep reinforcement learning,'' \emph{Sensors}, vol.~19, no.~18, p. 3837, 2019.

\bibitem{ref_1.4}
X.~Lei, Z.~Zhang, and P.~Dong, ``Dynamic path planning of unknown environment based on deep reinforcement learning,'' \emph{Journal of Robotics}, vol. 2018, 2018.

\bibitem{ref_1.5}
S.-Y. Chen, Y.~Yu, Q.~Da, J.~Tan, H.-K. Huang, and H.-H. Tang, ``Stabilizing reinforcement learning in dynamic environment with application to online recommendation,'' in \emph{Proceedings of the 24th ACM SIGKDD International Conference on Knowledge Discovery \& Data Mining}, 2018, pp. 1187--1196.

\bibitem{Ref_3-1}
\BIBentryALTinterwordspacing
V.~Radhakrishna, G.~R. Kumar, P.~V. Kumar, and V.~Janaki, ``A machine learning approach for imputation and anomaly detection in iot environment,'' \emph{Expert Syst. J. Knowl. Eng.}, vol.~37, no.~5, 2020. [Online]. Available: \url{https://doi.org/10.1111/exsy.12556}
\BIBentrySTDinterwordspacing

\bibitem{Ref_3-2}
\BIBentryALTinterwordspacing
B.~Yong, W.~Wei, K.~Li, J.~Shen, Q.~Zhou, M.~Wozniak, D.~Polap, and R.~Damasevicius, ``Ensemble machine learning approaches for webshell detection in internet of things environments,'' \emph{Trans. Emerg. Telecommun. Technol.}, vol.~33, no.~6, 2022. [Online]. Available: \url{https://doi.org/10.1002/ett.4085}
\BIBentrySTDinterwordspacing

\bibitem{ref_3-4}
I.~Cviti{\'c}, D.~Perakovi{\'c}, M.~Peri{\v{s}}a, and B.~Gupta, ``Ensemble machine learning approach for classification of iot devices in smart home,'' \emph{International Journal of Machine Learning and Cybernetics}, vol.~12, no.~11, pp. 3179--3202, 2021.

\bibitem{ref_3-6}
E.~Adi, A.~Anwar, Z.~Baig, and S.~Zeadally, ``Machine learning and data analytics for the iot,'' \emph{Neural computing and applications}, vol.~32, pp. 16\,205--16\,233, 2020.

\bibitem{ref_3-7}
M.~Ruta, F.~Scioscia, G.~Loseto, A.~Pinto, and E.~Di~Sciascio, ``Machine learning in the internet of things: A semantic-enhanced approach,'' \emph{Semantic Web}, vol.~10, no.~1, pp. 183--204, 2019.

\bibitem{ref_}
\BIBentryALTinterwordspacing
E.~Parisotto, F.~Song, J.~Rae, R.~Pascanu, C.~Gulcehre, S.~Jayakumar, M.~Jaderberg, R.~L. Kaufman, A.~Clark, S.~Noury, M.~Botvinick, N.~Heess, and R.~Hadsell, ``Stabilizing transformers for reinforcement learning,'' in \emph{Proceedings of the 37th International Conference on Machine Learning}, ser. Proceedings of Machine Learning Research, H.~D. III and A.~Singh, Eds., vol. 119.\hskip 1em plus 0.5em minus 0.4em\relax PMLR, 2020, pp. 7487--7498. [Online]. Available: \url{https://proceedings.mlr.press/v119/parisotto20a.html}
\BIBentrySTDinterwordspacing

\bibitem{ref__}
\BIBentryALTinterwordspacing
L.~C. Melo, ``Transformers are meta-reinforcement learners,'' in \emph{Proceedings of the 39th International Conference on Machine Learning}, ser. Proceedings of Machine Learning Research, K.~Chaudhuri, S.~Jegelka, L.~Song, C.~Szepesvari, G.~Niu, and S.~Sabato, Eds., vol. 162.\hskip 1em plus 0.5em minus 0.4em\relax PMLR, 2022, pp. 15\,340--15\,359. [Online]. Available: \url{https://proceedings.mlr.press/v162/melo22a.html}
\BIBentrySTDinterwordspacing

\bibitem{ref_2-1}
Z.~Wu, Z.~Liu, J.~Lin, Y.~Lin, and S.~Han, ``Lite transformer with long-short range attention.'' \emph{arXiv preprint arXiv:2004.11886}, 2020.

\bibitem{ref_2-2}
R.~Kozik, M.~Pawlicki, and M.~Chora{\'s}, ``A new method of hybrid time window embedding with transformer-based traffic data classification in iot-networked environment,'' \emph{Pattern Analysis and Applications}, vol.~24, no.~4, pp. 1441--1449, 2021.

\bibitem{rjoub2020bigtrustscheduling}
G.~Rjoub, J.~Bentahar, and O.~A. Wahab, ``Bigtrustscheduling: Trust-aware big data task scheduling approach in cloud computing environments,'' \emph{Future Generation Computer Systems}, vol. 110, pp. 1079--1097, 2020.

\bibitem{rjoub2023explainable}
{Rjoub, Gaith and Bentahar, Jamal and Wahab, Omar Abdel}, ``Explainable trust-aware selection of autonomous vehicles using lime for one-shot federated learning,'' in \emph{2023 International Wireless Communications and Mobile Computing (IWCMC)}.\hskip 1em plus 0.5em minus 0.4em\relax IEEE, 2023, pp. 524--529.

\end{thebibliography}

\end{document}